\documentclass[12pt]{article}
\usepackage{absspconf,amsmath,epsfig}
\makeatletter
\def\bstctlcite{\@ifnextchar[{\@bstctlcite}{\@bstctlcite[@auxout]}}
\def\@bstctlcite[#1]#2{\@bsphack
  \@for\@citeb:=#2\do{%
    \edef\@citeb{\expandafter\@firstofone\@citeb}%
    \if@filesw\immediate\write\csname #1\endcsname{\string\citation{\@citeb}}\fi}%
  \@esphack}
\makeatother 
\usepackage{enumitem}
\usepackage{booktabs}
\usepackage{color}
\usepackage[table]{xcolor}
\usepackage{float}
\usepackage{cite}

\title{Knowledge Distillation for Road Detection based on cross-model Semi-Supervised Learning}
\name{Wanli Ma, Oktay Karakuş, Paul L. Rosin}
\address{School of Computer Science and Informatics, Cardiff University, Cardiff CF24 4AG, UK}
\begin{document}
%
\maketitle

\begin{abstract}
The advancement of knowledge distillation has played a crucial role in enabling the transfer of knowledge from larger teacher models to smaller and more efficient student models, and is particularly beneficial for online and resource-constrained applications. The effectiveness of the student model heavily relies on the quality of the distilled knowledge received from the teacher. Given the accessibility of unlabelled remote sensing data, semi-supervised learning has become a prevalent strategy for enhancing model performance. However, relying solely on semi-supervised learning with smaller models may be insufficient due to their limited capacity for feature extraction. This limitation restricts their ability to exploit training data. To address this issue, we propose an integrated approach that combines knowledge distillation and semi-supervised learning methods. This hybrid approach leverages the robust capabilities of large models to effectively utilise large unlabelled data whilst subsequently providing the small student model with rich and informative features for enhancement. The proposed semi-supervised learning-based knowledge distillation (SSLKD) approach demonstrates a notable improvement in the performance of the student model, in the application of road segmentation surpassing the effectiveness of traditional semi-supervised learning methods.
\end{abstract}
%


\section{Introduction}
\label{sec:intro}

\bstctlcite{IEEEexample:BSTcontrol}

Road detection plays an important role in urban planning, navigation and routing, traffic management, emergency response, environmental impact assessment, and infrastructure maintenance \cite{wang2016review, bacher2005automatic, grote2007road, wei2017road}. In recent years, deep learning in computer vision has provided an efficient way for automated road detection and has become a common segmentation task in the literature. However, manually annotating large-scale and high-resolution remote sensing images is exceptionally time-consuming and expensive. One solution is to use semi-supervised learning approaches, which enable deep learning networks to learn from a small amount of labelled data together with a large amount of unlabelled data, thereby eliminating the labour-intensive and expensive annotation stage. Exploring additional unlabelled data through proper semi-supervised learning approaches offers significant benefits in training deep learning networks. These methods can attain competitive performance, compared even to those achieved by fully supervised learning approaches \cite{hu2021semi, zhang2021flexmatch, 9883979, 9875010, 9645575}. 

A significant challenge of (near-) real-time segmentation tasks like road surface detection is the trade-off between the accuracy of the segmentation and the speed at which it can be executed. More accurate methods generally require more model parameters and higher running time for inference. Whilst employing small models often leads to quicker task execution, they may lack the capacity to efficiently grasp and represent knowledge when provided with comparable computational resources and data as larger models. To solve this problem, knowledge distillation is used to transfer knowledge from large models to smaller ones without loss of validity \cite{hinton2015distilling}. However, nowadays, the majority of knowledge distillation studies in the field of remote sensing primarily focus on employing labeled data for training both teacher and student models, while not sufficiently investigating the use of unlabeled data to enhance the performance of student models.

This work proposes a semi-supervised learning-based knowledge distillation (SSLKD) method to train an efficient and lightweight machine learning model. Whilst leveraging an extensive pool of unlabelled data, the objective is to augment the variety of training samples for instructing both the teacher and the lightweight student models. This approach is crucial as the student model possesses limited capabilities in feature extraction, despite access to a substantial amount of unlabelled data. Simultaneously, knowledge distillation is employed to empower the lightweight student model to capitalise on the abundant features inherent in the more complex teacher models. 

Specifically, the contributions of this work are as follows: (1) We introduce a knowledge distillation process and a framework comprising two distinct teacher models to enhance the effectiveness of a lightweight and fast student model designed for road segmentation. Essentially, the framework undertakes extensive knowledge distillation across various levels, encompassing features, probabilities, and labels. Moreover, this distilled knowledge can be consistently applied to diverse segmentation tasks. 
(2) We employ a combination of semi-supervised learning and knowledge distillation methodologies. This strategy involves leveraging a significant volume of unlabelled data to enhance the capabilities of the lightweight student model. Simultaneously, the larger teacher models contribute comprehensive and informative guidance for the process of knowledge distillation.

\section{Related Work}
\label{sec:related_work}
Automatic road segmentation has been widely explored in the literature for decades \cite{lu2019multi, mattyus2017deeproadmapper, mnih2010learning, singh2018self}. Also, as the task of semantic segmentation has seen significant advancements in the computer vision field in recent years, a variety of 
deep learning models have been introduced in the literature, such as UNet \cite{ronneberger2015u}, SegNet \cite{badrinarayanan2017segnet}, PSPNet \cite{zhao2017pyramid}, DeepLabV3+ \cite{chen2018encoder}. These networks not only offer efficiency for pixel-wise road detection but also act as fundamental frameworks that researchers can refer to and modify to create networks specifically optimised for road detection \cite{zhu2021global, gao2018end}. In semi-supervised segmentation within the computer vision domain, DeepLabV3+ has emerged as one of the most used segmentation networks. This prominence is attributed to its advanced architecture, which combines the strengths of spatial pyramid pooling and encoder-decoder structures.

As a classic semi-supervised learning approach, consistency regularisation forces networks to give consistent predictions for unlabelled inputs that undergo diverse perturbations. Cross-consistency training (CCT) \cite{ouali2020semi} employs an encoder-decoder architecture with multiple auxiliary decoders where the consistency loss for unlabelled data is defined by mean-square-error (MSE) loss between predictions of the main decoder and the auxiliary decoders. Apart from introducing perturbations by using auxiliary decoders, network perturbation is used by guided collaborative training (GCT) \cite{ke2020guided} and cross pseudo supervision (CPS) \cite{chen2021semi} for consistency regularisation. Specifically, each of the aforementioned models uses two identical network structures but with different weight initialisation. CPS forces consistency by using pseudo labels, which are generated from network predictions, to mutually supervise the networks, whereas GCT achieves consistency regularisation by utilising predictions of networks cooperating with a flaw detector. CPS and CPS-to-$n$-networks (n-CPS) \cite{filipiak2021n} show considerable success with network perturbation on semi-supervised segmentation, which yields state-of-the-art semantic segmentation for benchmark datasets such as Cityscapes. However, CPS and n-CPS still restrict the diversity of pseudo labels and tend to output similar predictions since both of them use the same network architecture to generate pseudo labels. To increase the diversity of pseudo labels along with consistency regularisation, CGSSL \cite{ma2023confidence} and CNN\&Trans \cite{luo2022semi} use different segmentation networks to create pseudo labels and reach better segmentation performance for remote sensing and medical images, respectively.

Knowledge distillation employs a process of transferring rich and useful knowledge from a complicated teacher model to a smaller and faster student model \cite{hinton2015distilling}. Since the features and outputs from large models generally carry abundant and instructive information, they are regarded as soft labels to guide the supervision of a student. Feature-based knowledge distillation uses feature maps as knowledge sources for regular students \cite{heo2019comprehensive}. Nevertheless, when there is a substantial disparity in the model structures, identifying the specific areas where the features of the student and teacher demonstrate significant correlation, and can be efficiently employed for knowledge distillation, becomes a challenging task. To address this issue, the solution involves employing probability-based knowledge distillation, as proposed in studies such as Zhou et al. \cite{zhou2021rethinking} and Liu et al. \cite{liu2019structured}. In this approach, the knowledge distilled from models relies on their outputs, where irrespective of the model's structure, the elements of the n-dimensional output are ranged to [0,1] and sum to 1 and represent the probabilities associated with each class. The probability distribution takes different forms based on the nature of the tasks at hand. In the context of segmentation, it operates at a pixel-wise level, whereas for image classification, it is established at an image-wise level. Nevertheless, conventional knowledge distillation methods do not take into account the potential enhancement of both the teacher and student capabilities through incorporating unlabelled data.

\begin{figure}[h]
    \centering
    \includegraphics[width=0.7\linewidth]{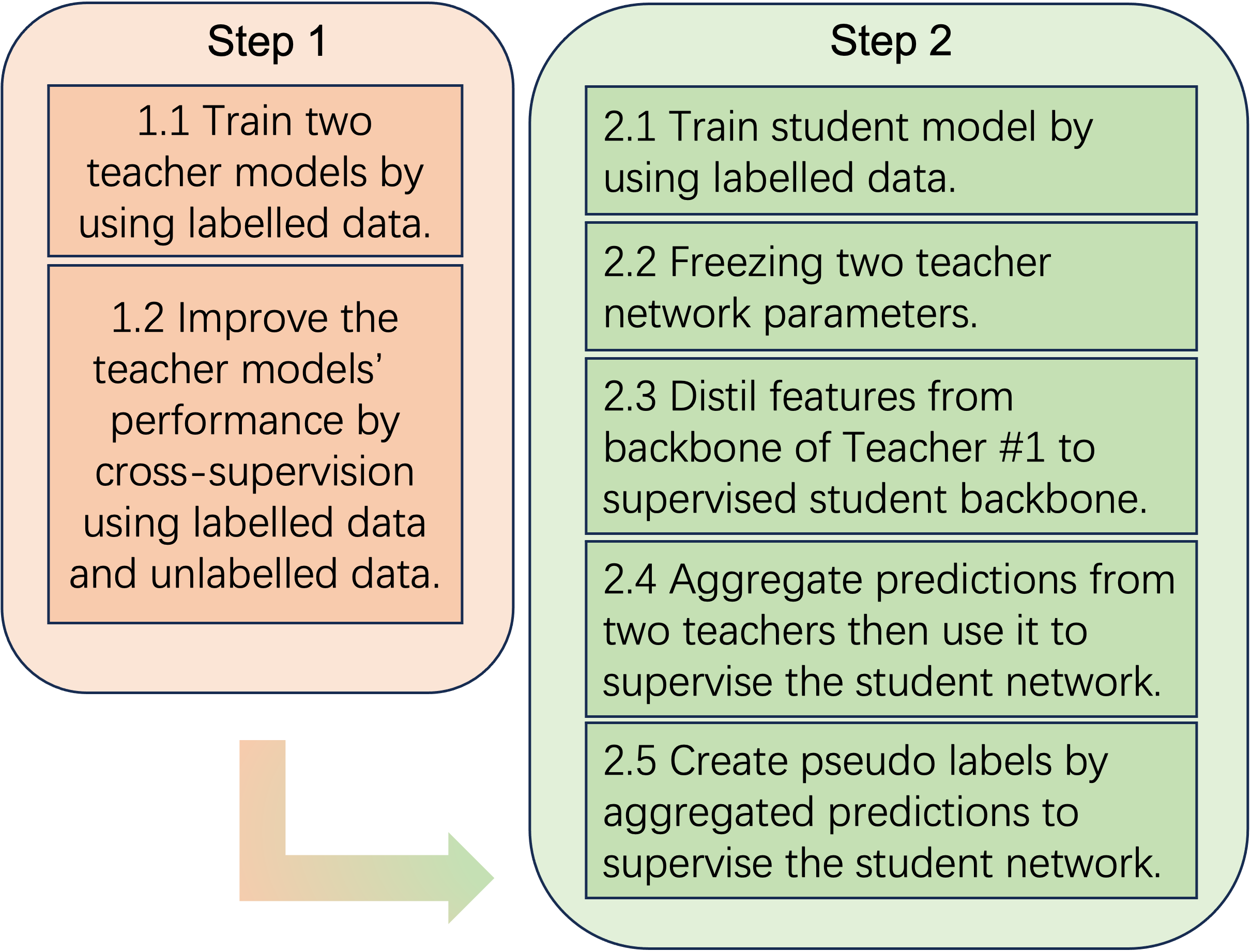}
    \caption{Steps of the Knowledge Distillation Procedure.}
    \label{fig:procedure}
    \vspace{-0.65cm}
\end{figure}

\section{method}
\label{sec:method}

\begin{figure*}[t]
    \centering
    \includegraphics[width=1\linewidth]{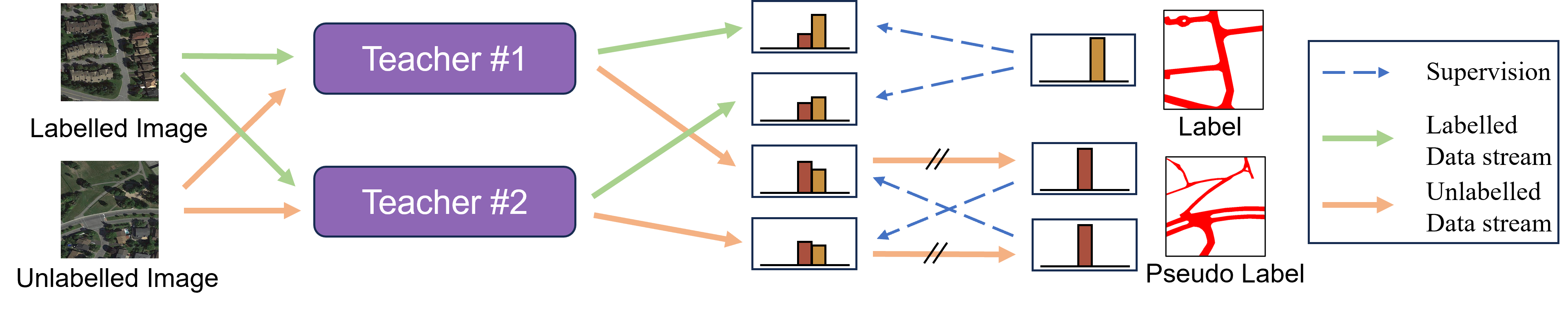}
    \caption{Framework of Cross-model Supervision.}
    \label{fig:Cross-model_Supervision}
    \vspace{-10pt}
\end{figure*}

\begin{figure*}[t]
    \centering
    \includegraphics[width=0.87\linewidth]{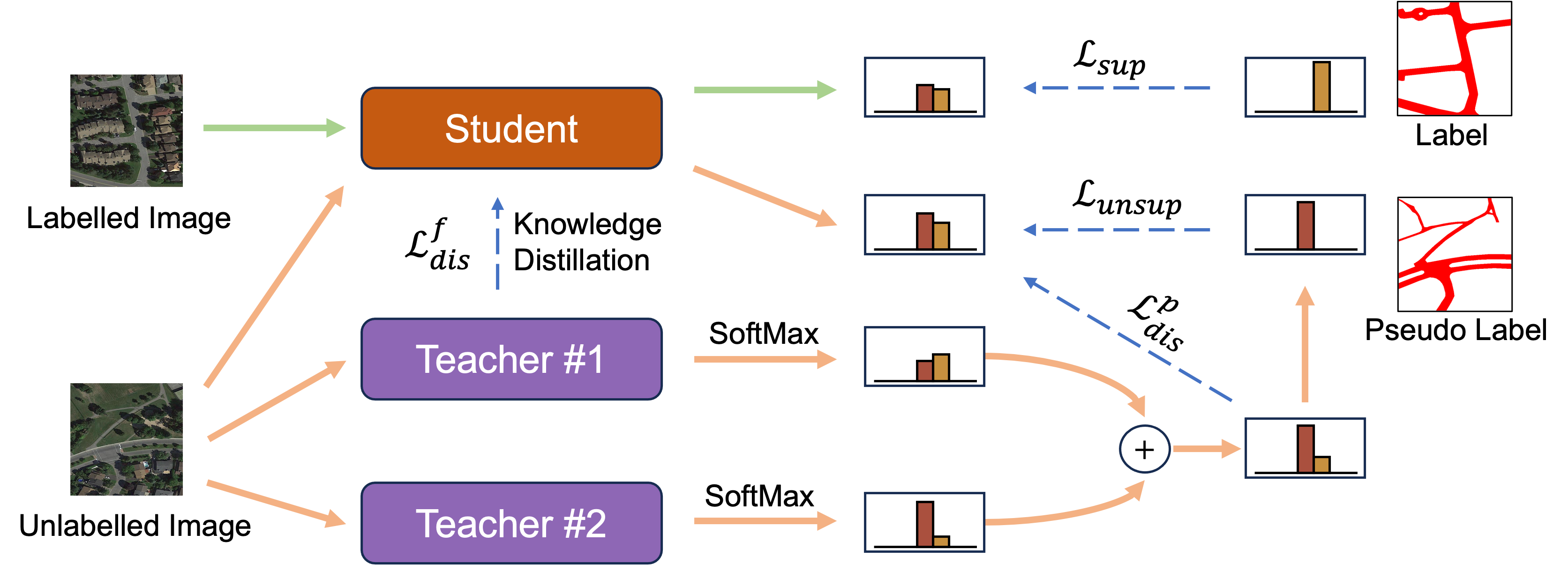}
    \caption{Framework of the proposed semi-supervised learning based knowledge distillation (SSLKD).}
    \label{fig:knowledge_distillation}
\end{figure*}

The proposed knowledge distillation includes two steps, as illustrated in Table \ref{fig:procedure}. In the first step, a small number of labelled data is used in a regular supervised learning manner to train the multiple teacher models by using the supervised loss $\mathcal{L}_{sup}$ between the $k^{th}$ ground truth $y_k$ and its corresponding prediction $p_k$. The number of teacher models can be more than two. However, due to GPU memory limitations, we have set the number of teacher models to two. The loss $\mathcal{L}_{sup}$ is defined as

\begin{equation}
\label{equ:equ1}
\mathcal{L}_{sup} = \frac{1}{W \times H} \sum_{k=1}^{W \times H}\ell_{c e}\left(p_{k}, y_{k}\right),
\end{equation}
where $W$ and $H$ refer to the weight and height of the input images whilst $\ell_{ce}$ is the standard cross-entropy loss function. 

Following this, to further improve the two teacher models’ representation to provide better features and pseudo labels for knowledge distillation, we implemented cross-model supervision to train the teacher models by using a large amount of unlabelled data. This is due to the fact that different networks typically possess complementary features and predictions, and they are regarded as teachers for one another. The framework of cross-model supervision is shown in Figure \ref{fig:Cross-model_Supervision} where Teacher \#1 refers to a DeepLabV3+ \cite{chen2018encoder} architecture with a Resnet-101 backbone, and Teacher \#2 represents a SegNet \cite{badrinarayanan2017segnet} with a VGG-16 backbone. Specifically, the predictions made by both teachers are employed to create pseudo labels, which are then utilised in training the other network through the cross-entropy loss function; however, these labels are not employed in the process of self-training.

The 2$^{nd}$ step of the proposed knowledge distillation method focuses on training a lightweight student model via SSLKD which is depicted in Figure \ref{fig:knowledge_distillation}. The student model is initialised by a small number of labelled data following the regular supervised learning by using the cross-entropy loss (\ref{equ:equ1}). After a certain number of training iterations, the procedure of knowledge distillation starts. The backbone knowledge $f_i$ of teacher \#1 is used to guide the extraction of feature $\hat{f}_i$ from the backbone of the student by using the distillation loss $\mathcal{L}_{dis}^{f}$ defined by mean absolute error (MAE)
\begin{equation}
\label{equ:equ2}
\mathcal{L}_{dis}^{f} = \frac{1}{C \times W \times H} \sum_{k=1}^{C \times W \times H} |f_i - \hat{f}_i| ,
\end{equation}
where $C$, $W$ and $H$ refer to the channel, weight and height of features, respectively. 

The reason behind using backbone knowledge from Teacher \#1 lies in the fact that both Teacher \#1 and the student share an identical network framework, which is Deeplabv3+. This similarity leads to their feature structure being closely aligned. In particular, the difference between the teacher and student models is in their respective backbones: the teacher model employs a more complex ResNet101 structure, whereas the student model utilises a lighter, less complex ResNet50 backbone. In addition to transferring knowledge at the backbone stage, the predictions from the two teachers are aggregated and ranged to [0-1] to be the probability for each class, and are then used to supervise the student network through $\mathcal{L}_{dis}^{p}$ based on MAE. Additionally, the combined predictions play a crucial role in generating pseudo labels, contributing to the supervision of the student network through cross-entropy loss $\mathcal{L}_{unsup}$. The total loss is the linear addition of the losses mentioned above. It is important to note that the two teachers underwent training during Step 1 and remained fixed in Step 2.

\begin{table*}[t]\renewcommand{\arraystretch}{0.65}
  \centering
  \caption{Comparing the performance of individual student and teacher models, along with students trained using the proposed method and other techniques, on the RoadNet dataset.}
    \begin{tabular}{p{7.5cm}cccccc}
    \toprule
     & IoU & OA & Precision & Recall & $F_1$ & GFLOPs \\
    \midrule
    Teacher \#1 \cite{chen2018encoder} & 68.32\% & 95.40\% & 88.45\% & 90.15\% & 89.29\% & 19.82  \\
    Teacher \#2 \cite{badrinarayanan2017segnet} & 70.65\% & 95.82\% & 89.14\% & 91.37\% & 90.24\% & 40.17  \\
    Student \cite{chen2018encoder} & 66.53\% & 95.04\% & 88.02\% & 89.07\% & 88.54\% & 14.94\\
    \midrule\midrule
    Teacher \#1 w/ unlabelled data\cite{chen2018encoder} & 71.96\% & 95.96\% & 90.25\% & 91.15\% & 90.70\% & 19.82  \\
    Teacher \#2 w/ unlabelled data\cite{badrinarayanan2017segnet} & 73.47\% & 96.25\% & 90.50\% & 92.11\% & 91.30\% & 40.17  \\
    \midrule\midrule
    Student w/ unlabelled data (CMS) \cite{ma2023confidence} & 70.85\% &95.74\% & 90.12\% & 90.38\% & 90.25\% & 14.94  \\
    
    CPS \cite{chen2021semi} & 69.74\% & 95.66\% & 88.84\% & \textcolor[rgb]{ 1,  0,  0}{\textbf{90.95\%}} & 89.88\% & 14.94  \\
    CCT \cite{ouali2020semi} & 68.99\% & 95.45\% & 89.15\% & 89.91\% & 89.53\% & 14.94  \\
    \textbf{Student (SSLKD)} & \textcolor[rgb]{ 1,  0,  0}{\textbf{71.89\%}} & \textcolor[rgb]{ 1,  0,  0}{\textbf{95.91\%}} & \textcolor[rgb]{ 1,  0,  0}{\textbf{90.57\%}} & 90.73\% & \textcolor[rgb]{ 1,  0,  0}{\textbf{90.65\%}} & 14.94  \\
    \bottomrule
    \end{tabular}%
    \vspace{-6pt}
  \label{tab:results}%
\end{table*}%

\begin{figure*}
    \centering
    \includegraphics[width=\linewidth]{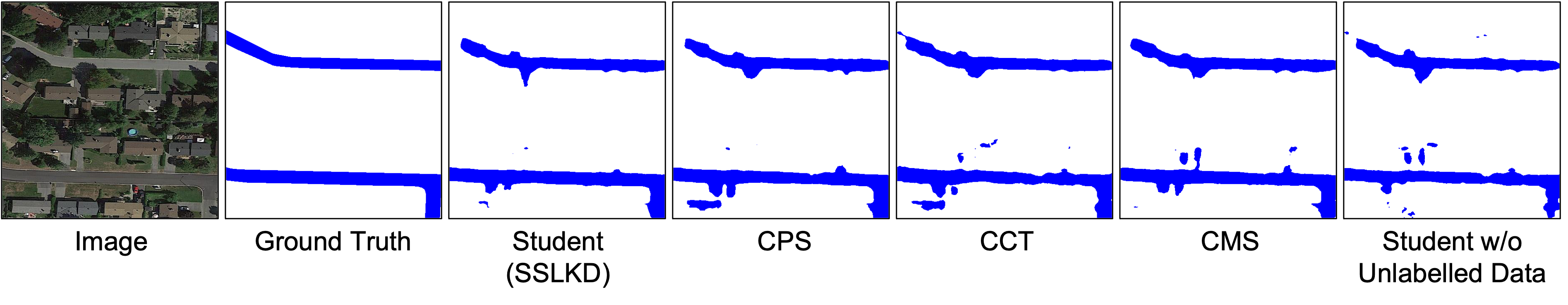}
    \caption{Visual segmentation results of the student model in each method on the RoadNet dataset.}
    \vspace{-10pt}
    \label{fig:Visual_results}
\end{figure*}

\section{EXPERIMENTS AND RESULTS}
\label{sec:experiments_Results}

We evaluated our method using the RoadNet dataset \cite{liu2018roadnet}, which consists of (1) image patches of size of 512 $\times$ 512, covering 21 city regions approximately \(8 \text{ km}^2\) with a spatial resolution of 0.21 meters per pixel. (2) Their manual labels for \textit{road edge}, \textit{centre line} and \textit{surface segmentation}. The numbers of samples for training and validation are 455 and 387, respectively. The ratio of the number of labelled and unlabelled data in experiments is 1:4.  

Our experiments were implemented in Pytorch. We used a mini-batch SGD optimiser that adopted a polynomial learning rate policy. All the experiments were performed on an NVIDIA A100-sxm in a GW4 Isambard. We thoroughly evaluated all models using class-related performance metrics, including overall accuracy (OA), precision, recall, intersection over union (IoU), and $F_1$-score.

The performance of teacher and student models trained separately by supervised learning are shown in the top three lines of Table \ref{tab:results}. Both of the two teacher models show superior performance compared to the student model as expected. Notably, the intersection over union (IoU) for Teacher \#2 is approximately 4\% higher than that of the student model. However, the values of GFLOPs for the teacher models exceed that of the student model. There is a significant enhancement in their performance after implementing cross-model supervision with unlabelled data for both teacher models. For instance, Teacher \#1 exhibits a 3.64\% improvement in IoU (line 4 of Table \ref{tab:results}).

Since the proposed SSLKD explored both labelled and unlabelled data, we conducted a performance evaluation of the proposed method against several state-of-the-art semi-supervised learning approaches, such as Cross-Model Supervision (CMS) \cite{ma2023confidence}, CPS \cite{chen2021semi}, and CCT \cite{ouali2020semi}, using the same train and test data set. Although we used the best Teacher \#2 for cross-model semi-supervised learning to supervise the student in CMS, the student performance of SSLKD is still better as shown in Table~\ref{tab:results}, thanks to the proposed combination of semi-supervised learning and knowledge distillation. Figure \ref{fig:Visual_results} demonstrates an example of predictions for all methods where SSLKD is predominantly closer to the ground truth compared to the students in other methods. 

\section{conclusion}
\label{sec:conclusion}
In this study, we introduced a novel hybrid methodology that combines knowledge distillation with semi-supervised learning to address road segmentation in very-high-resolution remote sensing imagery. The distillation process involves three distinct phases: feature, probability, and label levels. Our approach represents an expansion of traditional knowledge distillation through integrating semi-supervised learning, making it adaptable for application in various knowledge distillation frameworks. The proposed approach significantly enhances the performance of the student model in road segmentation, surpassing the effectiveness of previously proposed semi-supervised learning methods. Due to significant differences in model structures between Teacher \#2 and the student, feature-level knowledge distillation was not applied from Teacher \#2 to the student. In future work, we will primarily concentrate on investigating techniques for transferring feature knowledge from the teacher to the student, even in the presence of structural disparities between them.



\begingroup
\bibliographystyle{IEEEtran}
\setlength{\baselineskip}{0mm}
\setlength{\itemsep}{0em}
\setlength{\parskip}{0em}
\bibliography{refs}
\endgroup

\end{document}